\documentclass[3p,a4paper]{elsarticle}
\usepackage[utf8]{inputenc}

\title{Physics-informed Neural Networks with Periodic Activation Functions for Solute Transport in Heterogeneous Porous Media}

\author[mythirdaddress]{Salah A Faroughi \corref{mycorrespondingauthor}}
\author[mythirdaddress]{Ramin Soltanmohammadi}
\author[mythirdaddress]{Pingki Datta}
\author[mythirdaddress]{Seyed Kourosh Mahjour}
\author[mylastaddress]{Shirko Faroughi}
\cortext[mycorrespondingauthor]{Corresponding author: salah.faroughi@txstate.edu}

\address[mythirdaddress]{Geo-Intelligence Laboratory, Ingram School of Engineering, Texas State University, San Marcos, Texas, 78666, USA}
\address[mylastaddress]{Department of Mechanical Engineering, School of Engineering, Urmia University of Technology, Urmia, Iran}

\date{\today}

\usepackage{mwe}

\usepackage{setspace} 
\usepackage{tabularx}
\usepackage{lineno,hyperref}
\modulolinenumbers[5]

\usepackage{amsmath}
\usepackage{bm}
\usepackage{amssymb}
\usepackage[]{amsmath}
\usepackage{upgreek}

\let\today\relax
\makeatletter
\def\ps@pprintTitle{%
    \let\@oddhead\@empty
    \let\@evenhead\@empty
    \def\@oddfoot{\footnotesize\itshape
         {Submitted preprint- Dec 2022} \hfill\today}%
    \let\@evenfoot\@oddfoot
    }
\makeatother

\usepackage{pgfplots}
\pgfplotsset{compat=1.5}
\usepgfplotslibrary{units}
\usetikzlibrary{spy}
\usetikzlibrary{pgfplots.groupplots}
\usepackage{soul,color}
\usepackage{epstopdf}
\usepackage{units}
\usepackage{lscape}
\usepackage{booktabs}
\usepackage{gensymb}
\usepackage{multirow}

\usepackage{graphicx}
\usepackage{caption}
\usepackage{subcaption}

\usepackage{soul,color}
\soulregister\cite7
\soulregister\ref7
\soulregister\pageref7

\DeclareMathOperator\erfc{erfc}

\usepackage{floatrow} 
\usepackage[flushleft]{threeparttable}
\usepackage{tikz}
\usepackage{placeins}
\usepackage{color}
\usepackage{hyperref}

\usepackage{stackengine}

\usepackage{array}
\usepackage{tabularx}
\usepackage{pdfpages}
\usepackage{natbib}
\usepackage{textcomp, gensymb}

\begin{document}

\begin{abstract}
\color{black}
  Simulating solute transport in heterogeneous porous media poses computational challenges due to the high-resolution meshing required for traditional solvers. To overcome these challenges, this study explores a mesh-free method based on deep learning to accelerate solute transport simulation. We employ Physics-informed Neural Networks (PiNN) with a periodic activation function to solve solute transport problems in both homogeneous and heterogeneous porous media governed by the advection-dispersion equation. Unlike traditional neural networks that rely on large training datasets, PiNNs use strong-form mathematical models to constrain the network in the training phase and simultaneously solve for multiple dependent or independent field variables, such as pressure and solute concentration fields. To demonstrate the effectiveness of using PiNNs with a periodic activation function to resolve solute transport in porous media,  we construct PiNNs using two activation functions, \textit{sin} and \textit{tanh}, for seven case studies, including 1D and 2D scenarios. The accuracy of the PiNNs' predictions is then evaluated using absolute point error and mean square error metrics and compared to the ground truth solutions obtained analytically or numerically. Our results demonstrate that the PiNN with \textit{sin} activation function, compared to \textit{tanh} activation function, is up to two orders of magnitude more accurate and up to two times faster to train, especially in heterogeneous porous media. Moreover, PiNN's simultaneous predictions of pressure and concentration fields can reduce computational expenses in terms of  inference time by three orders of magnitude compared to FEM simulations for two-dimensional cases. 
\end{abstract}

\begin{keyword}
    Physics-informed Neural Networks\sep%
    Solute Transport\sep%
    Heterogeneous Porous Media\sep%
    Advection-Dispersion Equation\sep%
    Deep Learning\sep%
    Scientific Computing 
\end{keyword}

\maketitle

\color{black}
\section{Introduction}\label{sec:Intro}

Solute transport in porous media is crucial in many environmental and reservoir engineering applications, including risk and safety assessment of groundwater contamination \cite{cook2000determining}, secondary recovery processes in petroleum reservoirs \cite{gaus2008geochemical}, geological storage of radioactive waste \cite{pruess2002fluid}, hydrogen \cite{bienert2006membrane},  and carbon dioxide \cite{kristensen1999transport}, just to name a few \cite{li2020dfn}. It is a complicated process due to the varied temporal and spatial transport characteristics of the flow network \cite{hasan2020direct}. These characteristics can be divided into flowing areas influenced primarily by advection and stagnant regions governed mainly by dispersion \cite{faraji2022mathematical}. The process is considerably more complicated in heterogeneous porous media because all available pore space does not contribute to the flow uniformly; certain locations are dead ends, or transport is extremely slow due to inadequate connection to the main flow channels \cite{yang2019ubiquity}. Despite a conceptual understanding of the significance of multi-component solute transport mechanisms in porous media in general, attempts to conduct reliable experimental investigations are hindered by several barriers, resulting in a dearth of published information. The major causes are the technical difficulty in creating flow conditions and pore space connections under regulated laboratory or in situ test circumstances \cite{zhao2011numerical}. Hence, a system of field equations, which are numerically represented by partial differential equations (PDEs), is practically the only tool to predict the behavior of such non-linear flows in interconnected systems. In solute transport studies, traditional solvers, such as finite element (FEM), finite difference (FDM), and finite volume methods (FVM), have been extensively used to solve the governing PDEs. For example, \citet{zhang2018one} suggested a 1D model based on FEM for multi-component solute transport in saturated soil. \citet{bagalkot2015effect} presented a 1D numerical evaluation of the FDM for multi-species radionuclide transport in a single horizontally coupled fracture-matrix system. \citet{mostaghimi2016numerical} and \citet{maheshwari20133} investigated the effect of pore structure heterogeneity on reactive transport using a 3D pore scale model based on FVM. Although most large-scale calculations still use FEM, FDM, or FVM, they require high-resolution meshing to describe the geometry of the domain being represented. Because the size of an elementary mesh is normally too large, discrete domain descriptions may also need the availability of efficient transport equations at the mesh scale and the consideration of unresolved subgrid-scale effects \cite{noetinger2016random}. Hence, new methodologies and algorithms, especially mesh-free methods based on deep learning, are required to accelerate numerical simulations of solute transport through porous media.
\color{black}
Deep Learning (DL) methods have the potential to offer mesh-free solvers and address some of the aforementioned challenges.  Although most applications use DL (e.g., neural networks) to solve a lack of effective data modeling processes, explore vast design domains, and identify multidimensional connections \cite{im2021surrogate, karniadakis2021physics}, there is growing interest in using neural networks to solve PDEs \cite{raissi2019physics, han2018solving, berg2018unified, sirignano2018dgm}. In general, there are several main neural network frameworks to augment scientific computing: Physics-guided Neural Networks (PgNN),  Physics-informed Neural Networks (PiNN),  Physics-encoded Neural Networks (PeNN), and Neural Operators (NOs). The readers are referred to a recent review by \citet{faroughi2022physics}, where different neural network frameworks are compared head-to-head, and their challenges and limitations are thoroughly discussed. In this study, we adopt PiNN because of its straightforward mechanism to integrate the underlying physics compared to other approaches. By combining a loss function composed of the residuals of the physics equations, initial conditions, and boundary restrictions, PiNN-based models adhere to the physical laws. They employ automated differentiation to differentiate the output of neural networks with respect to their input (that is, spatio-temporal coordinates and model parameters) \cite{van2018automatic}. \color{black} The network can  estimate the solution with high precision by minimizing the loss function \cite{baydin2018automatic}. Therefore, PiNN provides the foundation for a mesh-free solver that incorporates long-standing advances in mathematical physics into DL \cite{raissi2019physics}.

\citet{raissi2017physics} developed PiNN as a new computing paradigm for forward and inverse modeling in a series of studies \cite{raissi2019physics, raissi2017physics, raissi2018hidden}.  \citet{raissi2018hidden} developed a PiNN framework, dubbed hidden fluid mechanics (HFM), to encode the physical laws governing fluid flows, i.e., the Navier-Stokes equations. They leveraged underlying conservation laws to derive hidden quantities of objective functions such as velocity and pressure fields from spatiotemporal visualizations of a passive scalar concentration in arbitrarily complex domains. Their technique accurately predicted 2D and 3D pressure and velocity fields in benchmark problems inspired by real-world applications. Since then, PiNN and its different variants have been applied aggressively to different fields, including porous media flows.  \citet{almajid2022prediction} employed the PiNN framework to solve both the forward and inverse problems of the Buckley-Leverett PDE equation representing two-phase flow in porous media. To test their implementation, they applied the classic problem of gas drainage through a porous water-filled medium. Several cases were examined to demonstrate the significance of the connectivity between observable data and PiNNs for various parameter spaces. According to their results, PiNNs are capable of capturing the solution's broad trend even without observed data, but their precision and accuracy improve significantly with observed data. \citet{hanna2022residual} employed PiNN to simulate one-dimensional (1D) and two-dimensional (2D) two-phase flow in porous media based on a new residual-based adaptive algorithm. Applying the PDE residual to build a probability density function from which additional collocation points are taken and added to the training set was fundamental to their work. The approach was applied individually to each PDE in the coupled system, considering the different collocation points for each PDE. Furthermore, the method was applied to enrich the points used to capture the initial and boundary conditions. They claimed that their approach yielded superior results with less generalization error than conventional PiNN with fixed collocation points. \citet{haghighat2022physics} introduced a PiNN method to solve coupled flow and deformation equations in porous media for single-phase and multiphase flow. Due to the problem's dynamic nature, they reported a dimensionless form of the coupled governing equations for the optimizer. 
In addition, they presented a way for sequential training based on the stress-split algorithms of poromechanics. They demonstrated that sequential training based on stress-split performs well for a variety of problems, whereas the conventional strain-split algorithm exhibits instability comparable to that observed for FEM solvers. \citet{he2020physics} extended a PiNN-based parameter estimation method to integrate multiphysics measurement. They studied a subsurface transport problem with sparse conductivity, a hydraulic head, and solute concentration. In their methodology, they employed the Darcy and advection-dispersion equations in conjunction with the data to train deep neural networks, reflecting space-dependent conductivity, head, and concentration fields. They proved that the proposed PiNN method considerably increased the accuracy of parameter and state estimates for sparse data compared to conventional deep neural networks trained with data alone. \citet{he2021physics} also suggested a discretization-free technique based on  PiNN for solving coupled advection-dispersion equations and the Darcy flow equation with space-dependent hydraulic conductivity. They used PiNN for 1D and 2D forward advection-dispersion equations and compared its performance for various Peclet numbers (Pe) with analytical and numerical solutions. They found that PiNN was accurate with errors of less than 1\% and outperformed other discretization-based approaches for large Pe. In addition, they proved that PiNN remained accurate for the backward advection-dispersion equations, with relative errors  below 5\% in the majority of instances. In another study, \citet{vadyala2022physics} implemented PiNN with the use of a machine learning framework such that it can be employed in reduced-order models to reduce the epistemic (model-form) ambiguity associated with the advection equation. They demonstrated that PiNN provided an accurate and consistent approximation with PDEs. Furthermore, they showed that PiNN could transform the physics simulation field by enabling real-time physics simulation and geometry optimization on large supercomputers.

\color{black}
Although there has been an increase in the application of PiNN to fluid flow studies in porous media \cite{rodriguez2022physics, zhang2022monotonicity, fuks2020limitations, zhang2023physics}, the application of PiNN to the advection–dispersion equation in heterogeneous porous media has received less attention due to the difficulty in predicting the velocity field imposed by the variation in the permeability field. The main objective of this study is to develop a PiNN model capable of accurately predicting transient solute transport in heterogeneous porous media. Due to the presence of  the second-order derivative in the governing equations,  selecting the activation function for PiNN becomes a critical step \cite{jagtap2022important}. \color{black} \textcolor{black}{To this end, a PiNN is constructed using periodic activation functions, and its convergence rate and accuracy are compared against a PiNN using \textit{tanh} to model transport phenomena in porous media}. Although it is theoretically believed that neural networks with periodic activation functions, e.g., sinusoids, may be harder to train, it is shown in several cases that periodicity is intuitively beneficial and can learn faster and better than other activation functions \cite{parascandolo2016taming,jagtap2022important}. In this study, using 1D and 2D test cases, we also show that a PiNN with a sine activation function provides more accurate predictions for solute transport in heterogeneous porous media while reducing the training time.

\color{black}

This paper is structured as follows: In Section~\ref{Solutetransport}, the underlying physics for solute transport in a porous medium is presented. Section \ref{PiNN} discusses PiNN's algorithm with periodic activation functions to predict solute transport. In Section~\ref{cases}, the PiNN is deployed to several one-dimensional (1D) and two-dimensional (2D) case studies, and its predictions are examined by comparing them with analytical and/or FEM solutions. Finally, in Section~\ref{sec:conclusion}, we summarize the main conclusions of this work.

\section{Underlying Physics}
\label{Solutetransport}
Solute transport in porous media  is governed by three main mechanisms: advection, diffusion, and dispersion   \cite{zhang2021influence, khan2022numerical, zhao1994solute}. The mathematical formulation of the advection-diffusion-dispersion equation is explained briefly here in terms of implementing the flow and transport factors that affect the transport process. The mass conservation equation serves as the basis for the solute  transport equation \cite{van1982analytical, sun2020review} that reads as,
\begin{equation}\label{mass transport}
    \nabla \cdot \mathbf{J} =  -\frac{\partial}{\partial{t}}(\phi{C}),
\end{equation}
where $\mathbf{J} ~[ML^{-2}T^{-1}]$ is the solute mass flux, $\phi$ is the porosity, and $C ~[ML^{-3}]$ is the solute concentration. 
The flux $\mathbf{J}$ contains two mechanisms,
\begin{equation}\label{solute mass flux}
    \mathbf{J} =  -\phi{D}\nabla{C} + \mathbf{u} (\phi{C}),
\end{equation}
where  the first term refers to  the diffusion flux, and the second term refers to the advection flux \cite{haigh2021eddy, lou2020effective}. In Eq.~(\ref{solute mass flux}), $D~[L^2T^{-1}]$ is the molecular diffusion coefficient ($D = D_0$), and $\mathbf{u}~[LT^{-1}]$ is the velocity vector of the pore fluid flow computed using Darcy's law,
\begin{equation}\label{Eq:velocity}
    \mathbf{u} = \mathbf{q}/\phi =  -\frac{k}{\mu\phi} (\nabla P + \rho \mathbf{g}),
\end{equation}
\textcolor{black}{where $\mathbf{q} ~[LT^{-1}]$ is the Darcy's flux, $P ~[ML^{-1}T^{-2}]$ refers to the pressure field, $\mathbf{g}~[LT^{-2}]$ is the gravity, $k ~[L^2]$ is the permeability, and $\mu ~[ML^{-1}T^{-1}]$ and $\rho~[ML^{-3}]$ are the fluid viscosity and density, respectively. To include the impact of  dispersion, the diffusion coefficient in Eq.~\ref{solute mass flux} can be changed to} 
\begin{equation}\label{dispersion}
    D =  D_0 + \alpha{U},
\end{equation}
\color{black} where $\alpha$ is known as the dynamic dispersivity, $D$ is now called the hydrodynamic dispersion coefficient, and $U$ is the magnitude of fluid velocity defined as $U = |\mathbf{u}|$. In this work, we assume that longitudinal and transverse dispersion are identical; however, we allow $U$ to vary spatially for a heterogeneous porous medium where $\mathbf{u}$ varies due to permeability changes \cite{talon2022statistical, baioni2021modeling}. The reason for this assumption is to focus on the impact of permeability  and how PiNN can handle it when supplied with different activation functions. \color{black} Substituting Eq.~(\ref{solute mass flux}) into Eq.~(\ref{mass transport}) leads to 
\begin{equation}\label{Eq:transportFinal}
    \nabla \cdot [\phi{D}\nabla{C}] - \nabla \cdot [\mathbf{u}(\phi{C})] =  \frac{\partial({\phi{C}})}{\partial{t}},
\end{equation}
which can be rewritten as,
\begin{equation}\label{Eq:transportFinal2}
    \nabla \cdot [\phi{D}\nabla{C}] - (\nabla \cdot \mathbf{u})[(\phi{C})] - \mathbf{u} \cdot \nabla(\phi{C}) =
    \frac{\partial({\phi{C}})}{\partial{t}},
\end{equation}
that incorporates diffusion, dispersion, and advection transports and serves as the foundation of solute transport in porous  media \cite{berkowitz2002physical}. If the solute transport process has no effect on fluid density  (i.e., incompressible flow, $\nabla.\mathbf{u} =0$), the velocity field can be calculated independently of the solute concentration using Darcy's law, and Eq.~(\ref{Eq:transportFinal2}) reduces to,
\begin{equation}\label{Eq:transportFinal3}
    \nabla \cdot [\phi{D}\nabla{C}] -  \mathbf{u} \cdot \nabla(\phi{C}) =
    \frac{\partial({\phi{C}})}{\partial{t}}.
\end{equation}

The flow incompressibility condition  generates an equation for computing the pressure field as,
\begin{equation}\label{pres}
 \nabla \cdot \mathbf{u} =  \nabla \cdot (-\frac{k}{\mu\phi} \nabla P + \rho \mathbf{g}) = 0,
\end{equation}
that reads as, 
\begin{equation}\label{pressure}
    \frac{\partial}{\partial{x}}[\zeta(x, y)\frac{\partial{P}}{\partial{x}}]+
    \frac{\partial}{\partial{y}}[\zeta(x, y)\frac{\partial{P}}{\partial{y}}] =0,
\end{equation}
for two-dimensional (2D) flow problems in x and y direction assuming $g_x = g_y= 0$, where
\begin{equation}\label{perm}
    \zeta(x, y)=\frac{k(x, y)}{\mu\phi}.
\end{equation}
 
 Equations (\ref{Eq:transportFinal3}) and  (\ref{pressure}) together form the governing equations for the concentration and pressure fields in a porous medium with a heterogeneous permeability distribution \cite{berkowitz2002physical}.

\section{Methodology}\label{PiNN}

\textcolor{black}{In this study, Physics-informed Neural Networks (PiNN)  \cite{raissi2019physics, karniadakis2021physics} are leveraged to resolve the
solute transport in porous media, which is governed by a strong mathematical form consisting of the pressure and advection-dispersion equations as well as the relevant initial and boundary conditions.} \textcolor{black}{Unlike Physics-guided Neural Networks (PgNNs), PiNN is a solution learning method that does not require labeled datasets \cite{faroughi2022physics}. In PiNNs, the underlying physics is included outside of the neural network architecture to constrain the model during training and  ensure that the outputs follow given physical laws. The most common method to emulate this process is through  a weakly imposed penalty loss that penalizes the network when it does not follow the physical constraints \cite{raissi2017physics, jagtap2020conservative, jagtap2021extended}}. 

\begin{figure}[htp]
    \centering
    \includegraphics[width=0.98\linewidth]{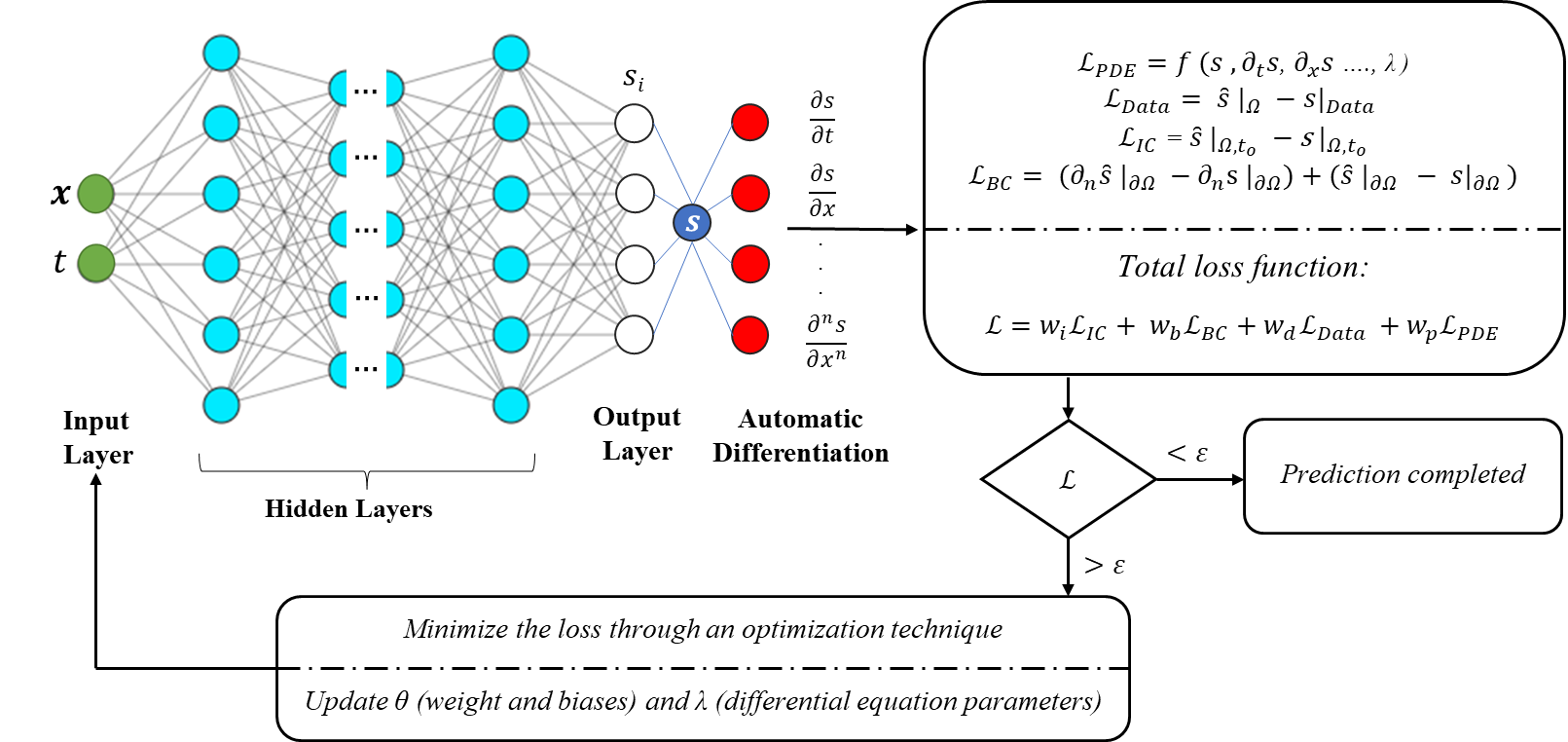}
    \caption{\textcolor{black}{A schematic architecture of Physics-informed Neural Networks (PiNNs). The network digests  spatiotemporal coordinates, (\textbf{\textit{x}},\textit{t}), as inputs to predict a solution set, $\textbf{\textit{s}}$, as an approximate to the ground truth solution, $\hat{\textbf{\textit{s}}}$. The automatic differentiation (AD) is then used to generate the derivatives of the predicted solution $\textbf{\textit{s}}$ with respect to inputs. These derivatives are used to formulate the residuals of the governing equations in the loss function weighted by different coefficients. $\theta$ and $\lambda$ are the learnable parameters for weights/biases and unknown PDE parameters, respectively, that can be learned  simultaneously while minimizing the loss function.}}
    \label{fig:PiNN}
\end{figure}

A schematic representation of our PiNN architecture is illustrated in Fig.~\ref{fig:PiNN}. As illustrated, a PiNN consists of three elements: a neural network (NN), an automatic differentiation (AD) layer, and a feedback mechanism informed by physics \cite{cuomo2022scientific}. A neural network is first built in order to digest the spatiotemporal features (i.e., $\textbf{x}$ and $t$) as input parameters and approximate the solution, $s(\mathbf{x},t)$, of a phenomenon described by PDEs with known boundary values. Assume that $\mathbb{N}_L{(\textbf{x})}:\mathbb{R}^{d_{in}}$ $\rightarrow$ $\mathbb{R}^{d_{out}}$ be a $L$-layer neural network, or a ($L-1$)-hidden layer neural network, with $N_l$ neurons in the $l^{th}$ layer ($N_0$ = $d_{in}$, $N_L$ = $d_{out}$) \cite{depina2022application}. Each layer in the NN is associated with its own weight matrix ${\textbf{W}^l}\in \mathbb{R}^{N_l \times N_{l-1}}$ and bias vector $\mathbf{b}^l \in \mathbb{R}^{N_l}$. The NN is recursively defined using an element-wise nonlinear activation function, $\sigma$, as follows, \cite{lu2021deepxde},
\color{black}
\begin{align}
   \textnormal{Input~layer:}~  \mathbb{N}_0(\mathbf{x}) &=  \mathbf{x} \in \mathbb{R}^{d_{in}}, \\
   \textnormal{Hidden~layers:}~    \mathbb{N}_l(\mathbf{x}) &=  \sigma({\textbf{W}^l} \mathbb{N}_{l-1}(\textbf{x})+\mathbf{b}^l) \in \mathbb{R}^{N_{l}}, ~~ for ~~1\le l \le L-1, \\
   \textnormal{Output~layer:}~     \mathbb{N}_L(\mathbf{x}) &=  {\textbf{W}^l} \mathbb{N}_{L-1}(\textbf{x}) + \mathbf{b}^l \in \mathbb{R}^{d_{out}},
\end{align}
where $\sigma$ is commonly specified as the logistic sigmoid $ \sigma (x) = 1/(1 - exp (-x))$, hyperbolic tangent $\sigma (x) = tanh~(x)$, or the rectified linear unit, $\sigma (x) = max~(x, 0)$ \cite{depina2022application}. In this work, we propose to use a sine (i.e., periodic) activation function,
\begin{equation}\label{hidden}
    \sigma({\textbf{W}^l} \mathbb{N}_{l-1}(\textbf{x})+\mathbf{b}^l) = \sin ({\textbf{W}^l} \mathbb{N}_{l-1}(\textbf{x})+\mathbf{b}^l),
\end{equation}
and compare its performance against \textit{tanh}, as one of the most used activation functions for PiNNs \cite{jagtap2022important}, in simulating solute transport in porous media. Periodic  activation functions  are ideally suited for representing complex physical signals and their derivatives  while converging faster than baseline architectures \cite{parascandolo2016taming,SIREN}. 
This superior behavior, in comparison to \textit{tanh}, is attributed to (i) the presence of simple sum and diff tasks for which a sine function can outperform hyperbolic tangent function \cite{parascandolo2016taming}, and (ii) the derivative of a network with sine activation function behaves like the network itself (i.e., the derivative of the sine is a cosine that is a phase-shifted sine) that potentially speeds up the learning process \cite{SIREN}, especially in  the solute transport problem that contains the second order derivatives in the governing equations.
\color{black}

The NN outputs are then fed into the  AD layer, which is the central property of PiNNs. AD is employed to assess the derivatives of the network outputs with respect to the network inputs \cite{rumelhart1986learning, depina2022application, van2018automatic, guo2020solving}. Consider a strong mathematical form  with a PDE specified in the domain $\Omega$ and parameterized by $\bm{\lambda}$, 
\begin{equation}\label{PDE}
   f \big(\textbf{x},t: \Delta s; \bm{\lambda} \big) = 0, ~~~ \mathbf{x}\in\Omega,
\end{equation}
and boundary conditions (e.g., Dirichlet, Neumann,  Robin boundary condition, etc.) specified on the boundary of the domain $\partial\Omega$,
\begin{equation}\label{bc}
    \psi (\mathbf{s}(\mathbf{x},t)) = 0, ~~~ \mathbf{x}\in\partial\Omega,
\end{equation}
where $\mathbf{s}$ is the unknown solution, and $\Delta$ represents the linear or nonlinear differential operator (e.g., $\frac{\partial}{\partial{t}}$, $\frac{\partial}{\partial{x}}$, $\frac{\partial^{2}}{\partial{x^{2}}}$, etc.). The initial condition can be considered as a kind of Dirichlet boundary condition in the spatiotemporal domain. AD is used to apply the $\Delta$ and $\psi$ operators to the neural network with respect to inputs, i.e.,  $\mathbf{x}$ and $t$, and  generate the required terms in the loss function in order to optimize the PDE solution, $\mathbf{s}$.  Lastly, a feedback mechanism is constructed to minimize the loss terms through optimizations. The total loss term is defined as \cite{cai2022physics,guo2020solving, strelow2023physics}, 
\begin{equation}\label{loss}
    \mathcal{L}={w_i}\mathcal{L}_{IC} + {w_b}\mathcal{L}_{BC} + {w_d}\mathcal{L}_{Data} + {w_p}\mathcal{L}_{PDE},
\end{equation}
where  $w_i$, $w_b$, $w_d$, and $w_p$, are referred to as the weights for the loss due to the initial conditions, boundary conditions, labeled data, if any, and PDEs, respectively. The individual loss terms are computed as,
\begin{align}
  \mathcal{L}_{IC} &= \frac{1}{N_{i}}\sum_{i=1}^{N_{i}} (\hat{s}{|_{\Omega,t_0}} - s{|_{\Omega,t_0}})^2,  \\
  \mathcal{L}_{BC} &= \frac{1}{N_{b}}\sum_{i=1}^{N_{b}}(({\partial_n}\hat{s}{|_{\partial{\Omega}}} - {\partial_n}s{|_{\partial{\Omega}}})-(\hat{s}{|_{\partial{\Omega}}} - s{|_{\partial{\Omega}}}))^2, \\
   \mathcal{L}_{Data} &= \frac{1}{N_{d}}\sum_{i=1}^{N_{d}}(\hat{s}{|_{\Omega}} - s{|_{Data}})^2, \\
   \mathcal{L}_{PDE} &= \frac{1}{N_{p}}\sum_{i=1}^{N_{p}} ( \emph{f}~({s}, ~{{\partial}_t}{s}, ~{{\partial}_x}{s}, ~\ldots,~\lambda))^2,
\end{align}
where $t_0$ is the initial time, $s$ is the neural network prediction, $\hat{s}$ is the ground truth solution, $N_i$, $N_b$, $N_d$, and $N_p$ are, respectively, the number of spatiotemporal points representing the initial conditions, boundary conditions, labeled data, and collocation points (i.e., spatiotemporal points within the domain where the neural network prediction, $s(\textbf{x},t)$ is checked against the constraints of PDEs). \color{black} PiNN with weights $\theta$ can then be trained by optimizing,
\begin{equation}
\label{eq:training}
    \theta^{\prime} = \arg \min_\theta \sum_{1}^{N^{*}} \mathcal{L}(\theta),
\end{equation}
leading to the PiNN with weights $\theta^{\prime}$ that generates reasonable predictions. $N^{*}$ represents the total number of input-output pairs in the training process. \textcolor{black}{PiNN training is more difficult than PgNN training, because PiNNs are composed of sophisticated non-convex and multi-objective loss functions, which may cause instability during optimization \cite{cuomo2022physics, shah2022physics}. The selection of weights for the loss terms is ad-hoc and problem (PDE) dependent. The weights are adjusted using trials and errors in the training phase to reach the minimum minimization error or mitigate the instability of the solution \cite{faroughi2022physics}. In this study, we  start with identical weights for all loss terms, and as needed (especially for heterogeneous cases where solutions encounter instability issues due to variation in the permeability field), we reduce the weight for PDE loss terms, $\omega_{p}$, to fully respect the boundary conditions and mitigate the instability in the solution. Also, we set $w_d =0$ as we do not provide labeled data to PiNN.}

Finally, a gradient-based optimizer such as Adam method \citep{kingma2014adam} and Limited-memory Broyden–Fletcher Goldfarb–Shanno with box constraints, L-BFGS-B \citep{byrd1995limited}, should be used to minimize the loss function. It is found that the PiNN predictions strongly depend on which of these two algorithms is selected \citep{lu2021deepxde}, and several studies suggested a two-step optimization algorithm that starts with the Adam method for a prescribed number of iterations, and then continues with the L-BFGS-B method until convergence \citep{he2020physics,he2021physics}. \textcolor{black}{In this work, L-BFGS-B, as a second-order, optimizer is used. It is observed that L-BFGS-B finds a satisfactory solution for smooth PDEs faster than Adam and with fewer iterations \cite{bengio2000gradient, kylasa2019gpu, richardson2018seismic}}. This is aligned with the goal of this work, which  compares PiNN's predictions when trained using different activation functions, without adding the complexity of alternating between optimizer schemes. Using L-BFGS-B,  at each iteration,  the loss is checked against $\mathcal{L} < \epsilon$, where $\epsilon$ is a specified tolerance. If the condition is not met, error backpropagation is implemented to update the learnable parameters ($\theta$ and/or $\lambda$). The entire cycle is repeated for a given number of iterations until the PiNN model produces learnable parameters with a loss error less than $\epsilon$ \cite{olmo2022physics}.

 \textcolor{black}{The proposed PiNN is deployed to model solute transport  in porous media by predicting pressure and concentration fields, i.e., $\textbf{s} = (P, C)$. We select the PiNN architectures using an iterative random-search hyperparameter tuning \cite{escapil2023hyper}  (e.g., selecting the number of layers,  neurons per layer, collocation points, and weights randomly within a specified range) for  different case studies in this work due to differences in spatial dimension,  boundary conditions, and permeability field. We ensure that the selected PiNN has enough layers and width to estimate each target field.} There are two critical normalization steps in the implementation of PiNNs that can be followed to ensure faster convergence to the correct solution \cite{rasht2022physics}, similar to the common practice of scaling and dimensionless analysis in other mesh-based computational techniques, e.g., FVM, FEM, etc. The first step is to map the network input and output variables to the interval $[0, 1]\in{R}$. The second step is to scale the pressure and concentration equations such that all terms are of the same order. In this work, we use the absolute point error (APE) and the mean squared error (MSE) defined as,
\begin{align}
    APE &= |s - \hat{s}|, \\
    MSE &= \frac{1}{N}\Sigma_{i=1}^{N} (s - \hat{s})^{2}, \label{eq: MSE} 
\end{align}
as statistical measurands to assess the accuracy of PiNNs' predictions, $s$, for a certain time-step against the ground truth solutions, $\hat{s}$,  obtained analytically or numerically using the finite element method. \color{black} Here, $N$ is the total number of inference points (resembling the spatiotemporal mesh) to generate the solutions for comparison.  

\color{black}

\section{Computational Experiments}\label{cases}

\color{black} In this section, the proposed PiNN model is applied to solve the 1D and 2D solute transport phenomena under different conditions (i.e., a total of seven computational experiments with homogeneous and heterogeneous domains). We compare the PiNN models with \textit{sin} and \textit{tanh} activation functions and validate them against analytical and/or FEM solutions to examine their capability and accuracy. The comparisons are presented in terms  of accuracy and training time. 

\color{black}

\subsection{Case 1: 1D solute transport with constant velocity}
This test case explores solute transport in a 1D  domain ($x\in [0,L]$) representing an isotropic porous medium  with a constant velocity field. As the steady state velocity field is known, the governing equation, Eq.~(\ref{Eq:transportFinal3}), reduced to,
\begin{equation}\label{1D advection–diffusion}
    \frac{\partial{C}}{\partial{t}} + {u_x} \frac{\partial{C}}{\partial{x}} = \frac{\partial}{\partial{x}}(D_x\frac{\partial{C}}{\partial{x}}), ~~~~ 0<x<L,~~~~ 0<t<t_0,
\end{equation}
where $L = 1$ is the length of the domain, $u_x = 0.5~m/s$ is the steady state velocity field, and $D_x = 0.02~m^2/s$ refers to the hydrodynamic dispersion coefficient in the $x$ direction. For this case, we assume the following initial and boundary conditions,
\begin{equation}\label{1D BC}
\begin{cases}
    C(x, t) = 0; ~~~ t = 0, \\
    C(x, t) = C_0; ~~~ x = 0, \\
    \frac{\partial{}}{\partial{x}} C(x,t) = 0; ~~ x = L.
\end{cases}
\end{equation}
where $C_0= 1.0 ~kg/m^3$ is the injected concentration at $x= 0$ (i.e., Dirichlet boundary condition). This 1D advection–dispersion can be solved analytically \cite{zhou2009lattice}. The analytical solution is,
\begin{equation}\label{analytical solution}
    C(x, t) = C_0 \left( 1 - 2\exp(\frac{xu_x}{2D_x} - \frac{{u_x^2}t}{4D_x}) \sum_{i=1}^{\infty}\frac{{\beta_i}\sin({\frac{\beta_i{x}}{L}})\exp(-\frac{{\beta_i^2}D_x{t}}{L^2})}{{\beta_i^2} + {({\frac{{u_x}L}{2D_x}})^2} + \frac{{u_x}L}{2D_x}} \right),
\end{equation}
where $\beta_i$ are the roots of,
\begin{equation}\label{roots}
    \beta \cot{\beta} + \frac{{u_x}L}{2D_x} = 0.
\end{equation}

Equations (\ref{analytical solution}) and (\ref{roots}) can be approximated using \cite{van1982analytical},
\begin{equation}\label{approximate equation}
    C(x, t) = C_0 ~ A(x, t), ~~ 0<t \leq t_0,
\end{equation}
where $A (x, t)$ is computed as,
\begin{equation}\label{A_1DSolution}
\begin{aligned}
    A(x, t) &= \frac{1}{2} \erfc\left(\frac{x-u_x{t}}{2\sqrt{{D_x}t}}\right)\;+\;\frac{1}{2}\exp(\frac{u_x{x}}{D_x}) \;\erfc \left(\frac{x+{u_x}t}{2\sqrt{{D_x}t}}\right) \\ 
    & + \frac{1}{2}\left(2 + \frac{{u_x}(2L-x)}{{D_x}} +  \frac{{u_x^2}t}{{D_x}} \right)\;
    \exp (\frac{{u_x}L}{D_x})\erfc \left(\frac{2L-x + {u_x}t}{2\sqrt{{D_x}t}}\right) \\ &- ({\frac{{u_x^2}t}{\pi{{D_x}}}})^{\frac{1}{2}}  \exp\left(\frac{{u_x}L}{D_x} - \frac{1}{4D_xt} (2L-x+{u_x}t)^2\right). 
\end{aligned}
\end{equation}

\begin{figure}[htp]
    \centering
    \includegraphics[width=0.99\linewidth]{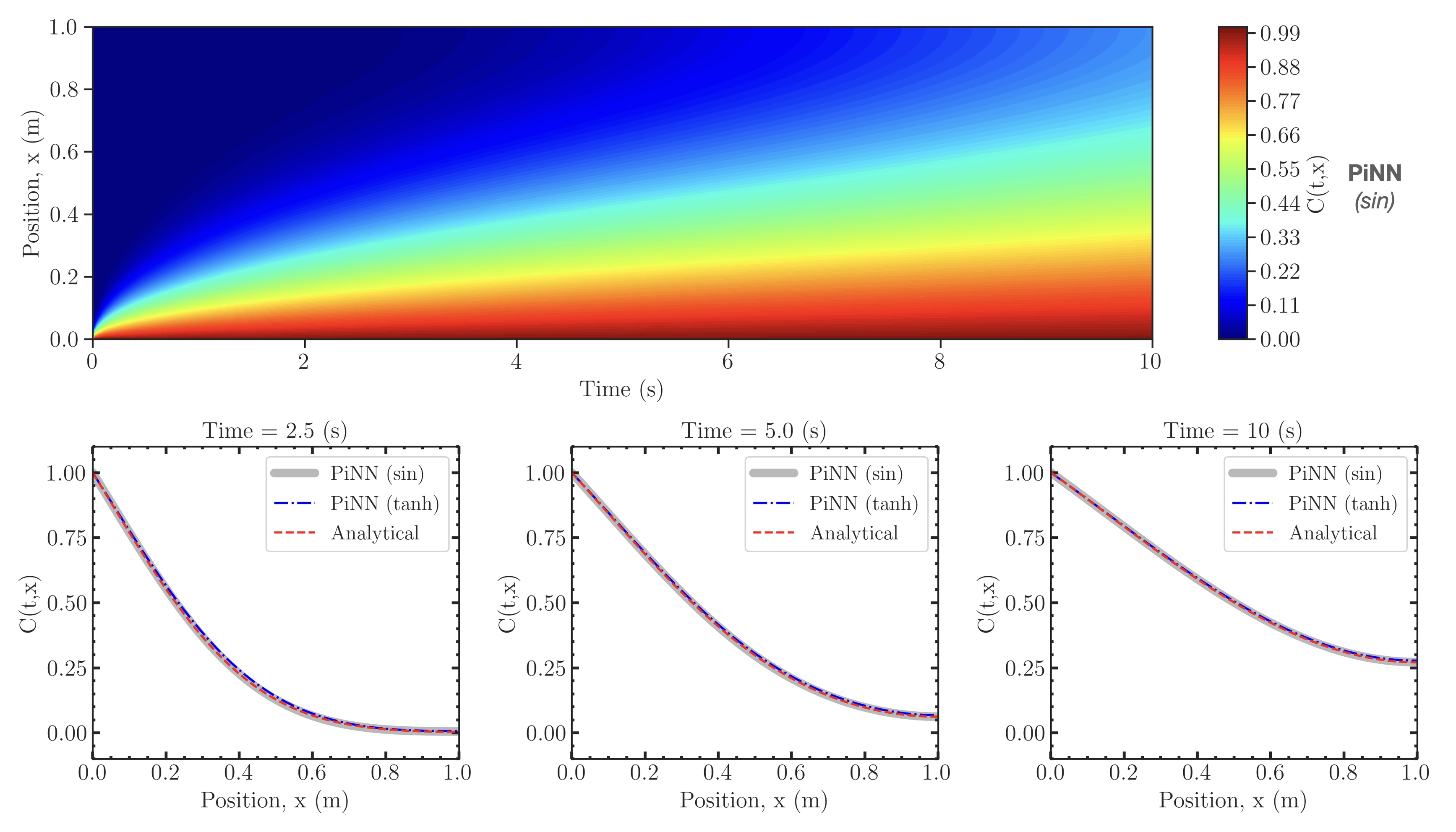}
    \caption{One-dimensional solute transport with a constant velocity field, $u_x = 0.5~m/s$. The upper panel shows the concentration field  predicted by PiNN with \textit{sin} activation function within the spatiotemporal domain $(0\leq x\leq1, ~0 \leq t \leq 10)$ at an injection rate of $C_0 = 1 kg/m^3$. The lower panels show the comparison of  PiNNs' prediction with the ground truth obtained analytically using Eq.~(\ref{approximate equation})   at three different times. The comparison demonstrates a good agreement yielding $MSE = 1.15 \times 10^{-6}$ using \textit{sin} activation function and $MSE = 1.21 \times 10^{-6}$ using \textit{tanh} activation function for the entire spatiotemporal domain.}
    \label{fig:Results 103_1D}
\end{figure}

\color{black}
A PiNN with \textit{sin} activation function and randomly distributed collocation points (with a uniform distribution scheme in the specified domain) is used to predict the solute transport in 1D domain described by Eq.~(\ref{1D advection–diffusion}) with the initial and boundary conditions given in Eq.~(\ref{1D BC}). An iterative random-search hyperparameter tuning process  is practiced to find the best architecture based on the MSE accuracy measures. This process yielded a network with four hidden layers with \{32, 32, 16, 16\} neurons per layer,  5,000 spatiotemporal random points within $0\leq x\leq1$ and $ 0 \leq t \leq 10$ as the collocation points to enforce the PDE loss term, 5,000 spatiotemporal random points to collectively represent the boundary and initial conditions informing the loss terms, and identical weights for all loss terms ($w_i = w_b = w_p = 1$). Another PiNN with similar architecture but using the \textit{tanh} activation function is also trained and tuned for comparison purposes, see Table \ref{tab:my-tableCase12}. For this PiNN model, the number of layers, neurons per layer, and collocation points are fixed, but the weights are allowed to change for better optimization. These assumptions make the comparisons fair in terms of both accuracy and training time.

Figure \ref{fig:Results 103_1D} represents the concentration field in the ($\mathbf{x}, t$) domain predicted by PiNN using \textit{sin} activation function. It also shows the PiNNs' prediction compared with the ground truth (analytical solution given by Eq.~(\ref{A_1DSolution}) at different simulation times. \textcolor{black}{The comparison demonstrates a good agreement yielding $MSE = 1.15 \times 10^{-6}$ using \textit{sin} activation function and $MSE = 1.21 \times 10^{-6}$ using \textit{tanh} activation function for the entire spatiotemporal domain.} As reported in Table \ref{tab:my-tableCase12}, for this 1D case, the PiNN with \textit{tanh} activation function approximates the solution with the same order of magnitude accuracy, but with nearly 25\% increase in training time captured based on three runs using a system with 3.00-GHz 48-core Intel Xeon Gold 6248R CPU, Nvidia Quadro RTX 8000 GPU, and 128 GB of RAM. In the next test cases, to better differentiate the models, the PiNNs are applied to more complex problems representing the complexity of solute transport in porous media.

\begin{table}[H]
\caption{A comparison of PiNN models using \textit{sin} and \textit{tanh} activation functions in terms of accuracy and training time to predict the concentration field in homogeneous 1D and 2D porous media, validated against the analytical solutions. The MSE is calculated using Eq.~\ref{eq: MSE}  in (t,x) domain for Case 1, and in (x,y) domain at $t=1.0~(s)$ for Case 2 assuming $\textbf{u} = (u_x, u_y)= (0.0, 0.50)~m/s$. The training time of the PiNNs is measured through three runs on a system equipped with a 3.00-GHz 48-core Intel Xeon Gold 6248R CPU, Nvidia Quadro RTX 8000 GPU, and 128 GB of RAM. Additionally, the number of collection points in the domain and the weights assigned to the loss terms are reported for each PiNN model.}
\label{tab:my-tableCase12}
\begin{tabular}{p{4.8cm}p{2cm}p{2cm}p{2cm}p{2cm}}
\toprule
& \multicolumn{2}{l}{\textbf{Case 1}} & \multicolumn{2}{l}{\textbf{Case 2}} \\   \cline{2-5}   \addlinespace
& \textit{sin}     & \textit{tanh}    & \textit{sin}     & \textit{tanh}    \\ \hline
 \addlinespace
Collocation Points, $N_p$ & 5000 & 5000 & 8000 & 8000      \\  \addlinespace
Weights: $w_i$, $w_b$, $w_p$ & 1.0, 1.0, 1.0 & 1.0, 1.0, 0.9 & 1.0, 1.0, 1.0 & 1.0, 1.0, 0.8     \\ \addlinespace
Accuracy (MSE) & \textbf{1.15e-06} &  1.21e-06 &   \textbf{1.54e-06} & 2.37e-05         \\  \addlinespace
Training Time (s)     & \boldsymbol{$86 \pm 7$}&  $108 \pm 3$&  \boldsymbol{ $388 \pm 24$}  & $509  \pm 15$                 \\ \bottomrule
\end{tabular}
\end{table}

\color{black}

\subsection{Case 2: 2D solute transport with constant velocity}

\begin{figure}[]
    \centering
    \includegraphics[width=0.99\linewidth]{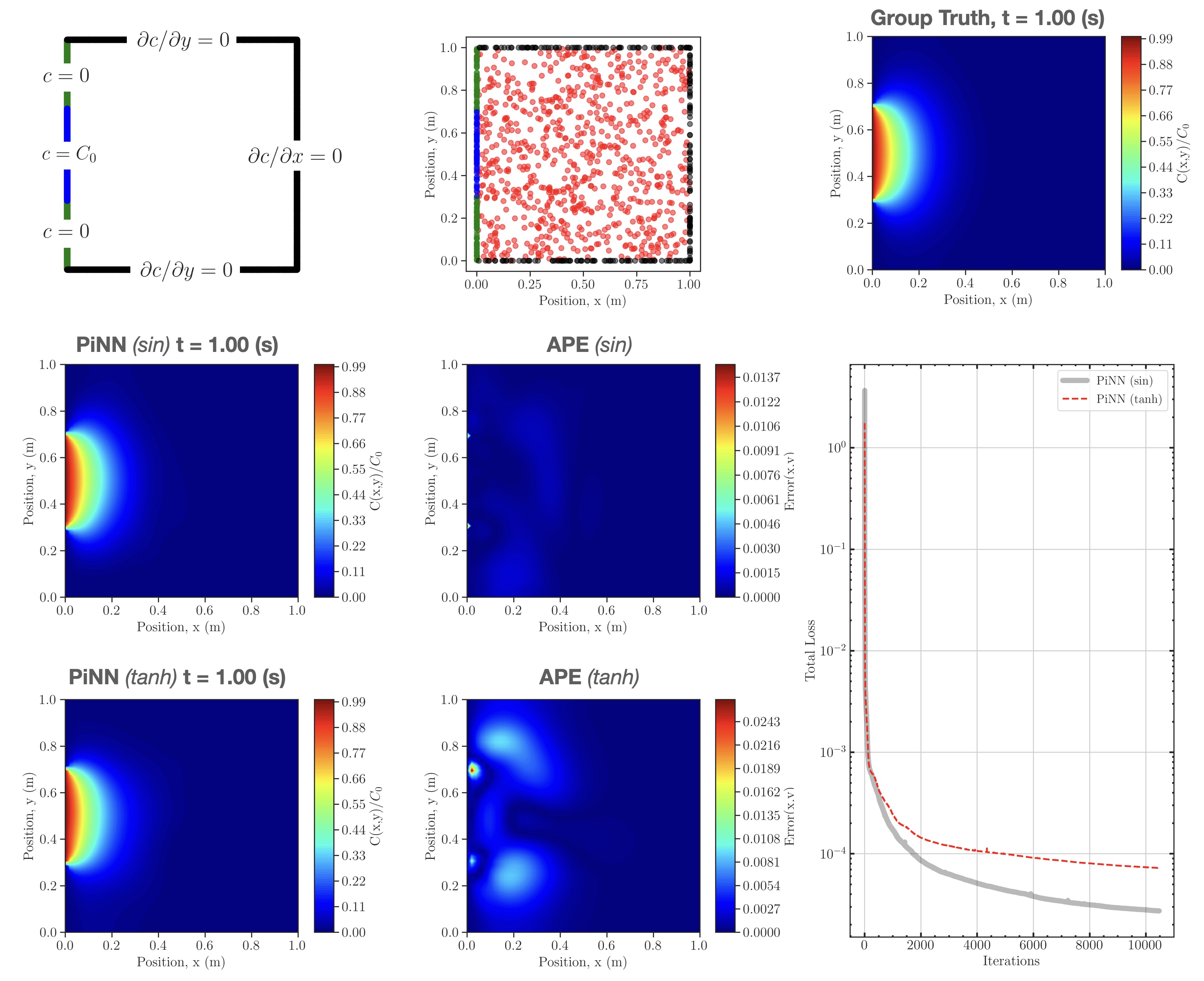}
    \caption{A comparison between the predictions of PiNNs with \textit{sin} and \textit{tanh} activation functions  and ground truth for solute transport in a 2D domain representing an isotropic porous medium  considering  $D_x = D_y = 0.02~m^2/s$, $\textbf{u} = (u_x, u_y)= (0.0, 0.0)~m/s$, and a solute injection rate of $C_0= 0.2 ~kg/m^3$ between $y_1 = 0.3~m$ and $y_2 = 0.7~m$. The upper panels show the domain with boundary conditions and distributions of randomly selected points on which different  terms in the loss function are evaluated. The red dots represent sample collocation points inside the domain corresponding to the loss term associated with the 2D solute transport PDE, and the blue, green, and black dots  represent the points on the boundary of the domain corresponding to the loss terms associated with the boundary conditions. The lower panel shows the comparison between the PiNNs' predictions for the concentration field at $t=1.00~s$ and the ground truth obtained analytically using Eq.~(\ref{analytical solution2D}). The absolute point error shows the mismatch between  the solutions, and the total loss plot depicts the difference in the convergence rate of PiNNs. }
    \label{fig:Results 103_2D_DIF}
\end{figure}

This test case explores solute transport in a 2D rectangular domain ($x\in [0,L],~y\in[0,W]$) representing an isotropic porous medium with a constant velocity field, $\textbf{u} = (u_x, u_y)$. Since the velocity field is given, the governing equation for solute transport, Eq.~(\ref{Eq:transportFinal3}), is simplified to,
\begin{equation}\label{2D advection–diffusion}
    \frac{\partial{C}}{\partial{t}} + {u_x} \frac{\partial{C}}{\partial{x}} + {u_y} \frac{\partial{C}}{\partial{y}} = \frac{\partial}{\partial{x}}(D_x\frac{\partial{C}}{\partial{x}}) + \frac{\partial}{\partial{y}}(D_y\frac{\partial{C}}{\partial{y}}), 
\end{equation}
where $D_y = D_x = 0.02~m^2/s$ refers to the hydrodynamic dispersion coefficient in the $y$ and $x$ directions. As schematically shown in Fig.~\ref{fig:Results 103_2D_DIF}, the following initial and boundary conditions,
\begin{equation}\label{BC2D}
\begin{cases}
    C(x,y, t) = 0; ~~~ t =0,  \\
    C(x,y, t) = C_0; ~~~ x = 0 ~~and ~~{y_1}\le{y}\le{y_2}, \\
    C(x,y, t) = 0; ~~~ x = 0 ~~and ~~y<{y_1} ~~and ~~y> {y_2}, \\
    \frac{\partial{}}{\partial{x}}C(x,y, t) = 0; ~~ x = L, ~~and~~ 0\le y\le W,\\
    \frac{\partial{}}{\partial{y}}C(x,y, t) = 0; ~~ y=0~~ or ~~y=W,  ~~and~~ 0\le x\le L,
\end{cases}
\end{equation}
are considered for this test case. The blue points on the left boundary bounded by $y_1$ and $y_2$ represent the positions of point-source solute injection at the rate of $C_0= 0.2 ~kg/m^3$. The other three sides of the domain are assigned as zero-gradient boundaries, assuming a free outflow of solute. This 2D advection-dispersion problem defined by Eqs.~(\ref{2D advection–diffusion}) and (\ref{BC2D}) can be solved analytically \cite{zhou2009lattice},
\begin{equation}\label{analytical solution2D}
    C(x, y, t) = C_0\sum_{n=0}^{\infty}{L_n}{P_n}\cos{({\eta}y)} \left( \exp 
    (\frac{x({u_x}-\zeta)}{2{D_x}}) \erfc (\frac{x-{\zeta}t}{2\sqrt{{D_x}t}}) + \exp(\frac{x({u_x}+\zeta)}{2{D_x}})\erfc(\frac{x+{\zeta}t}{2\sqrt{{D_x}t}})
    \right),
\end{equation}
where $L_n$ is defined as,
\begin{equation}\label{Ln}
  {L_n}=\begin{cases}
    \frac{1}{2}; ~~ n=0,\\
    1; ~~ n>0,
  \end{cases}
\end{equation}
and $P_n$ is computed as,
\begin{equation}\label{Pn}
  {P_n}=\begin{cases}
    ({y_2}-{y_1})/W; ~~ n=0,\\
    (\sin{(\eta{y_2})} - \sin{(\eta{y_1})})/(n\pi); ~~ n>0,
  \end{cases}
\end{equation}
where $\eta$ and $\zeta$, respectively, are,
\begin{equation}\label{eta}
  {\eta} = \frac{n{\pi}}{W} ~~ (n = 0,1,2,3,\ldots), ~~~~~~\zeta = \sqrt{{{u_x}^2} + 4{{\eta}^2}{D_x}{D_y}}.
\end{equation}

\textcolor{black}{First, a PiNN is employed with \textit{sin} activation function and randomly distributed collocation points} to predict the solute transport in the 2D domain ($L = W = 1~m$) considering the dispersion only with $D_x = D_y = 0.02~m^2/s$ and $\textbf{u} = (u_x, u_y)= (0.0, 0.0)~m/s$ in Eq.~(\ref{2D advection–diffusion}) and the initial and boundary conditions given by Eq.~(\ref{BC2D}). The solute injection rate is set to $C_0 = 0.2~kg/m^3$ between $y_1 = 0.3~m$ and $y_2 = 0.7~m$. To determine the best  architecture, similar to the previous case, an iterative random-search hyperparameter tuning approach is practiced. This approach produced a network with four hidden layers and \{32,16,16,16\} neurons per layer, 8,000 spatiotemporal random points within $0\leq x, y \leq1, ~0 \leq t \leq 1$ as collocation points to enforce the PDE loss term, 8,000 spatiotemporal random points to collectively represent the boundary and initial conditions informing the loss terms, and identical weights for all loss terms ($w_i = w_b = w_p = 1$).

Figure \ref{fig:Results 103_2D_DIF} shows the comparison of the 2D concentration fields predicted by PiNNs with \textit{sin} and \textit{tanh} activation functions and the analytical solution given by Eq.~(\ref{analytical solution2D}). The absolute point error is also shown, which illustrates a good agreement between the analytic solution and PiNN's prediction. Based on the total loss plot, it can be observed  that the PiNN with \textit{sin}  activation function is faster to converge and more accurate. The mismatch with the analytical solution in both cases is considerable at locations where extremely high concentration gradients exist (e.g., close to points $(0.0,0.30)$ and $(0.0,0.70)$). This mismatch is due to a difficult-to-minimize approximation error, i.e., PiNN struggles to converge to ground truth solutions close to those points due to the inherent complexity that arises in high-gradient areas and the limited capacity of the network architecture and training procedure \cite{yu2022gradient,chiu2022can}. A PiNN with a deeper network may resolve those areas and converge to the correct solution, but that makes minimization error a harder task.     

\color{black}

\begin{figure}[]
\centering\includegraphics[width=0.99\linewidth]{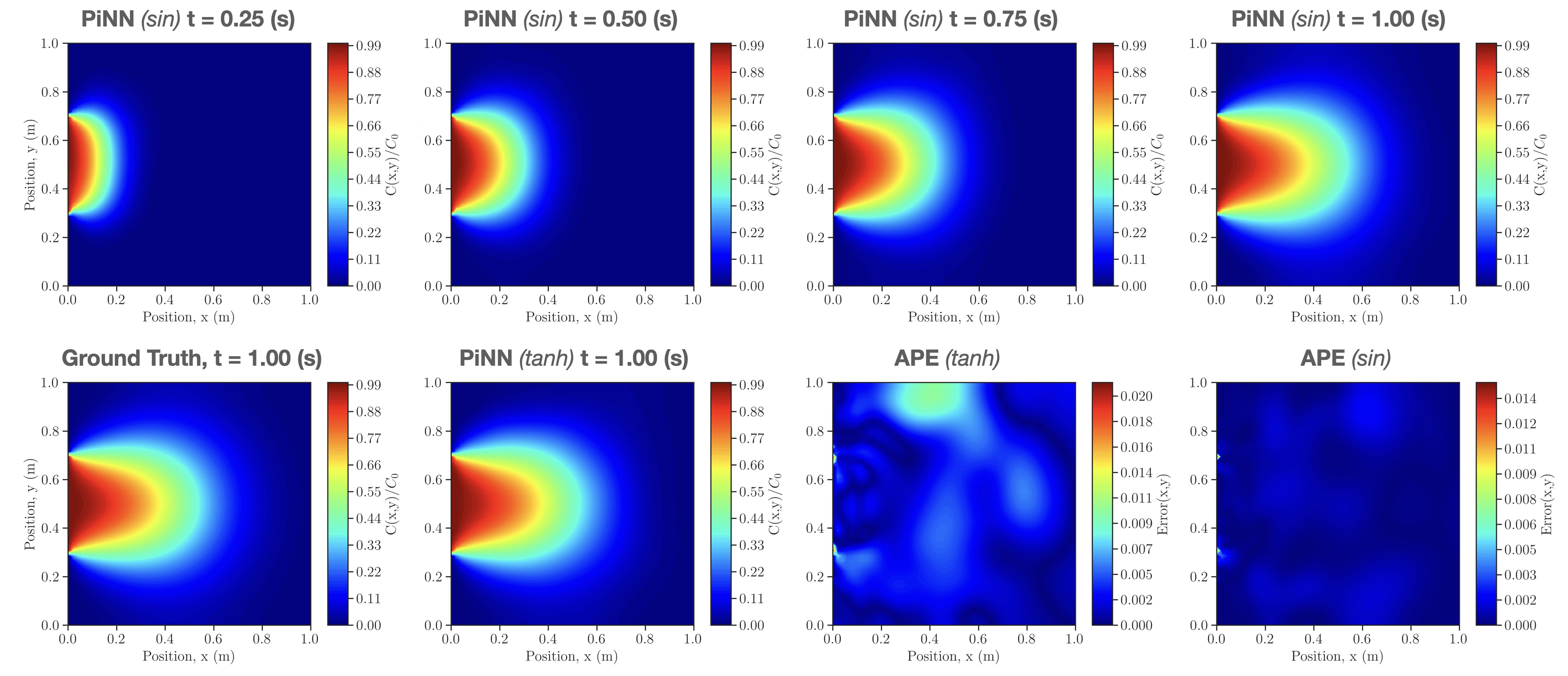}
    \caption{A comparison between the prediction of PiNN with \textit{sin} and \textit{tanh} activation functions and ground truth obtained analytically for the concentration field considering $D_x = D_y = 0.02~m^2/s$, $\textbf{u} = (u_x, u_y)= (0.5, 0.0)~m/s$, and an injection rate of $C_0= 0.2 ~kg/m^3$ between $y_1 = 0.3~m$ and $y_2 = 0.7~m$. The upper panels show the predictions of the PiNN with \textit{sin} activation function at $t=0.25~s,0.50~s,0.75~s$, and $1.00~s$. At $t=1.00~s$, the PiNNs' predictions are compared with the analytical solution obtained using Eq.~(\ref{analytical solution2D}). The absolute point error shows the mismatch between  the solutions yielding the MSEs reported in Table \ref{tab:my-tableCase12}.}
    \label{fig:Results 103_2D_ad}
\end{figure}

Two PiNN models with \textit{sin} and \textit{tanh} activation functions and the same architecture as discussed above are then employed to model the advection–dispersion solute transport in 2D domains with $D_x = D_y = 0.02~m^2/s$ and $\textbf{u} = (u_x, u_y)= (0.5, 0.0)~m/s$. The solute is injected again at a rate of $C_0 = 0.2 ~kg/m^3$ between $y_1 = 0.3~m$ and $y_2 = 0.7~m$. Table \ref{tab:my-tableCase12} reports the results of the comparison of these two models against the ground truth. The concentration fields  predicted by PiNN with \textit{sin} activation function at different times ($t=0.25~s,~0.50~s,~0.75~s,~1.00~s$) are shown in Fig.~\ref{fig:Results 103_2D_ad}.
The predictions of PiNNs for the concentration field at $t = 1.00~s$ are also compared with the analytical solution given by Eq.~(\ref{2D advection–diffusion}). The absolute point error shows that there is a good match between the PiNNs' predictions, with \textit{sin}, and the analytical solution within the entire domain yielding $MSE = 1.54 \times 10^{-6}$ that is about an order of magnitude more accurate compared to the PiNN with \textit{tanh} activation function while being nearly 31\% faster to be trained.

\subsection{Case 3: 2D solute transport in homogeneous porous media }\label{Case 3}

In this test case, the solute transport in a 2D  homogeneous porous medium ($x\in [0,L],~y\in[0,W]$) is investigated. For this problem, assuming $\zeta(x,y)= 1$, the pressure equation defined by Eq.~(\ref{pressure}) reduces to,
\begin{equation}\label{pressure2D}
    \frac{\partial^2{P}}{\partial{x}^2}+
    \frac{\partial^2{P}}{\partial{y}^2}=0,
\end{equation}
that must be solved to determine the velocity field using Eq.~(\ref{Eq:velocity}). The governing equation for the solute concentration remains the same as  Eq.~(\ref{2D advection–diffusion}). The following boundary conditions are considered for the pressure field,
\begin{equation}\label{BC2D_p}
\begin{cases}
    P(x, y) = 1.0; ~~ y=0~~ or~~y=W, ~~and ~~ 0\leq x \leq L,\\
    P(x, y) = 0.1; ~~ x=0~~ or~~x=L, ~~and ~~ 0\leq y \leq W,
\end{cases}
\end{equation}
and for the concentration field, 
\begin{equation}\label{BC2D_homo}
\begin{cases}
    C(x, y,t) = 0; ~~y=0~~ or~~y=W~~and~~x<{x_1}~~and~~ x>{x_2}, \\
    C(x, y,t) = C_0; ~~y=0~~ or~~y=W~~and~~{x_1}\le{x}\le{x_2}, \\
    \frac{\partial{}}{\partial{x}}C(x, y,t) = 0; ~~ x=0~~ or~~x=L, ~~ and ~~ 0\leq y \leq W, 
\end{cases}
\end{equation}
with $C(x, y, t= 0)= 0$ as the initial condition. The pressure value at the corner points that belong to two perpendicular sides is set to an average value of $0.55$, as additional constraints, to maintain symmetry in the solution. The ground truth solutions for pressure and concentration fields are obtained using FEM with $100\times100$ quadrilateral elements assuming $L = W = 1~m$, $D_x = D_y = 0.02~m^2/s$, and $C_0 = 0.2~kg/m^3$. 

\begin{figure}[H]
    \centering
    \includegraphics[width=0.99\linewidth]{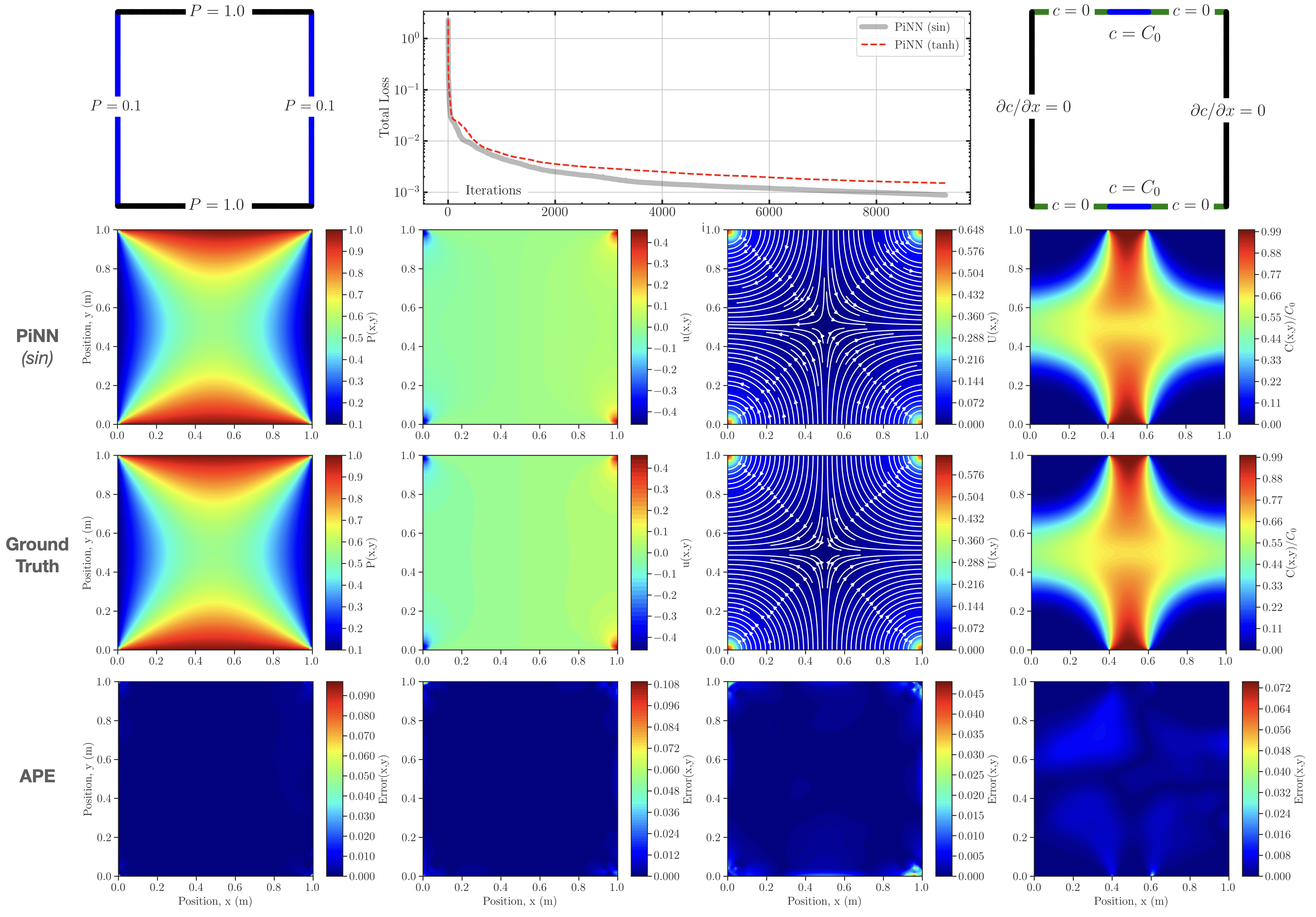}
    \caption{\textcolor{black}{A comparison between the prediction of the PiNN with \textit{sin} activation function and the ground truth for the pressure field in a 2D homogeneous porous medium considering $D_x = D_y = 0.02~m^2/s$ and $C_0 = 0.2~kg/m^3$.  The upper panels show the domain with pressure and concentration boundary conditions, as well as the comparison of total loss vs. iteration for PiNNs using \textit{sin} and \textit{tanh} activation functions. The lower panels show the comparison between the PiNN's predictions and ground truth solutions obtained using FEM with $100\times100$ quadrilateral elements for  the pressure field, the velocity field in the $x$ direction, and the total velocity field under steady state conditions as well as the concentration field at $t = 1.00~s$. The flow direction of the pore fluid  is also represented by the streamlines. The absolute point error is also shown for each field to illustrate the mismatch between the solutions.}}
    \label{fig:Results 103_2D_homo_p}
\end{figure}

A PiNN with \textit{sin} activation function and randomly distributed collocation points is employed to predict the solute transport in the 2D porous media described above. The PiNN's inputs are the spatiotemporal coordinates, $(x,y,t)$, and the outputs are the pressure and concentration fields, i.e.,  $\textbf{s} = \{P,C\}$ in Fig.~\ref{fig:PiNN}. The pressure and concentration fields are decoupled, therefore one may use two PiNNs side-by-side to solve this problem; however, training a PiNN with multiple outputs is more desirable in this study to minimize hyperparameter tuning. The iterative random-search hyperparameter tuning process is again practiced  to obtain the best PiNN architecture based on the MSE accuracy measure. This approach resulted in a network with five hidden layers and \{32, 16, 16, 8, 8\} neurons per layer using 12,000 spatiotemporal random points within $0\leq x,y \leq1, ~0 \leq t \leq 1$ as collocation points to enforce the pressure and concentration PDEs' loss terms, 12,000 spatiotemporal random points to collectively represent the boundary and initial conditions loss terms (sample selected points are shown in Fig.~\ref{fig:Results 103_2D_homo_p}), and identical weights for all loss terms ($w_i = w_b = w_p = 1$). Again, a PiNN with similar architecture but using \textit{tanh} activation function is also trained and tuned for comparison purposes, see Table \ref{tab:my-tableCase3}.

\begin{table}[]
\caption{A comparison of PiNN models using \textit{sin} and \textit{tanh} activation functions in terms of accuracy and training time to predict the pressure and concentration fields in homogeneous  2D porous media, validated against the ground truth solutions obtained using FEM with $100\times100$ quadrilateral elements. The MSE is calculated in (x,y) domain using Eq.~\ref{eq: MSE}  for the pressure field at steady state, and for the concentration field at $t=1.00~(s)$. The training time of the PiNNs is measured through three runs on a system equipped with a 3.00-GHz 48-core Intel Xeon Gold 6248R CPU, Nvidia Quadro RTX 8000 GPU, and 128 GB of RAM. Additionally, the number of collection points in the domain and the weights assigned to the loss terms are reported for each PiNN model.}
\label{tab:my-tableCase3}
\begin{tabular}{p{10.8cm}p{1.8cm}p{2cm}}
\toprule
& \multicolumn{2}{l}{\textbf{Case 3}}  \\   \cline{2-3}   \addlinespace
& \textit{sin}     & \textit{tanh}       \\ \hline
 \addlinespace
Collocation Points, $N_p$ & 12000 & 12000     \\  \addlinespace
Initial and Boundary Loss Weights: $w_i$, $w_b$  & 1.0, 1.0  & 1.0, 1.0    \\ \addlinespace
Pressure and Concentrations PDE Loss Weights: $w^P_p$, $w^C_p$  & 1.0, 1.0  & 0.7, 0.9    \\ \addlinespace
Pressure MSE &      \textbf{2.84e-06}  &  4.36e-05         \\  \addlinespace
Concentration MSE & \textbf{1.22e-06}  &  1.51e-05         \\  \addlinespace
Training Time (s)     & \boldsymbol{ $946 \pm 81$}  &  $1687 \pm 53$          \\ \bottomrule
\end{tabular}
\end{table}

Figure \ref{fig:Results 103_2D_homo_p} depicts a comparison between  ground truth (FEM solutions) and PiNN's predictions, with \textit{sin} activation function, for the pressure field, the velocity field in the $x$ direction, the total velocity field, and the flow streamlines in a homogeneous porous medium. The absolute point error is also displayed for all fields, indicating good agreement between PiNN and FEM solutions. 
Figure \ref{fig:Results 103_2D_homo_p} also demonstrates the comparison between the PiNN's prediction and FEM solution for the concentration field at $t=1.00s$. The absolute point error indicates the agreement between both solutions. The results reported in Table \ref{tab:my-tableCase3} reveal that the PiNN with \textit{sin} activation function is almost an order of magnitude more accurate than its counterpart (the PiNN with \textit{tan} activation function), while its training time is reduced by 78\%. Furthermore, the faster convergence of PiNN with the \textit{sin} activation function is noticeable in the total loss plot versus iteration, as illustrated in Fig.~\ref{fig:Results 103_2D_homo_p}. Based on these results, it can be inferred that the proposed PiNN is capable of simultaneously solving for the pressure and solute concentration fields in a 2D homogeneous porous medium with high accuracy. The next section of the study  explores the application of PiNN and the benefit of using a periodic activation function to solve a more complex problem of solute transport imposed by permeability heterogeneity in porous media.  

\color{black}

\subsection{Case 4: 2D solute transport in heterogeneous porous media }

\begin{figure}[H]
\centering
\includegraphics[width=0.99\linewidth]{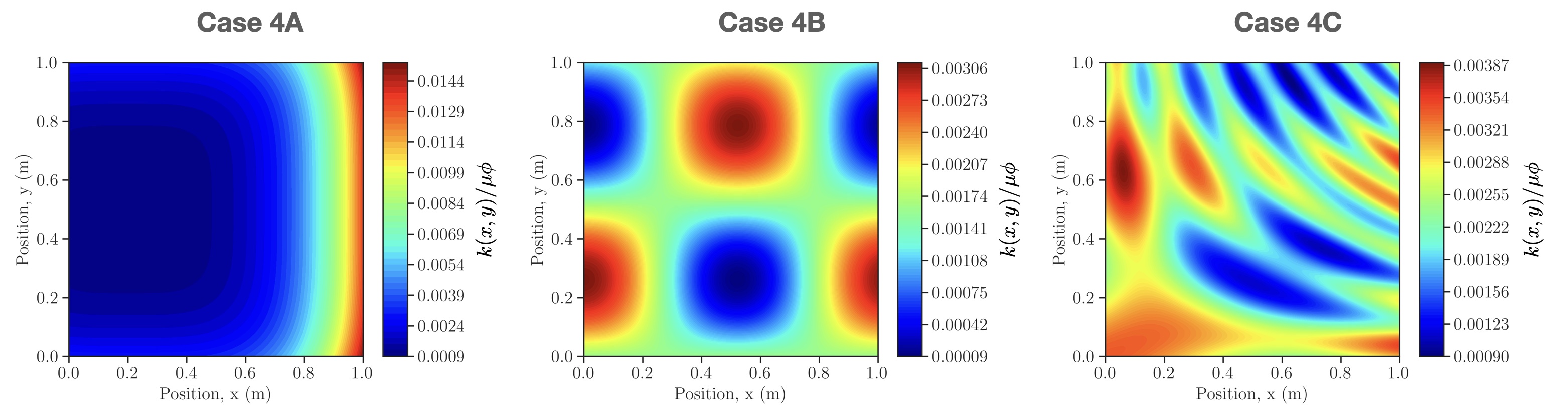}
    \caption{\textcolor{black}{Scaled permeability fields  selected to possess structural features with different length-scales compared to the domain size. From left to right, i.e., Case 4A, 4B, and 4C, the length-scale of the structural features decreases, leading to more intricate problems to simulate using PiNNs owing to the existence of high-frequency features. }}
    \label{fig:perms}
\end{figure}

In this test case, the solute transport is analyzed in a 2D rectangular domain ($x\in [0,L],~y\in[0,W]$) representing a heterogeneous porous medium, i.e., the permeability $k(x,y)$, and hence, $\zeta(x,y)$ in Eq.~(\ref{pressure}) vary spatially. \textcolor{black}{Three different $\zeta(x,y)$ fields, i.e., scaled permeability fields, are considered, as shown in Fig.~\ref{fig:perms}. 
These fields  are designed to exhibit structural features with distinct length scales relative to the size of the domain}. \textcolor{black}{As shown in Fig.~\ref{fig:perms}, from left to right, i.e., Case 4A, 4B, and 4C, the length scale of the structural features in the permeability field decreases, leading to more complex problems to simulate using PiNNs owing to the increase in high-frequency features.} The initial and boundary conditions for these cases are identical to those reported for Case 3 (refer to Section \ref{Case 3}). The pressure constraints at the corner points that correspond to the two perpendicular sides are not maintained due to variations in permeability. For each case, the ground truth solutions for the pressure and concentration fields  are determined using FEM with $100\times100$ quadrilateral elements considering $L = W = 1~m$, $\alpha = 0$, $D_x = D_y = 0.02~m^2/s$, $\mu \phi = 0.001$, and $C_0 = 0.2~kg/m^3$.

\begin{figure}[H]
\centering
\includegraphics[width=0.99\linewidth]{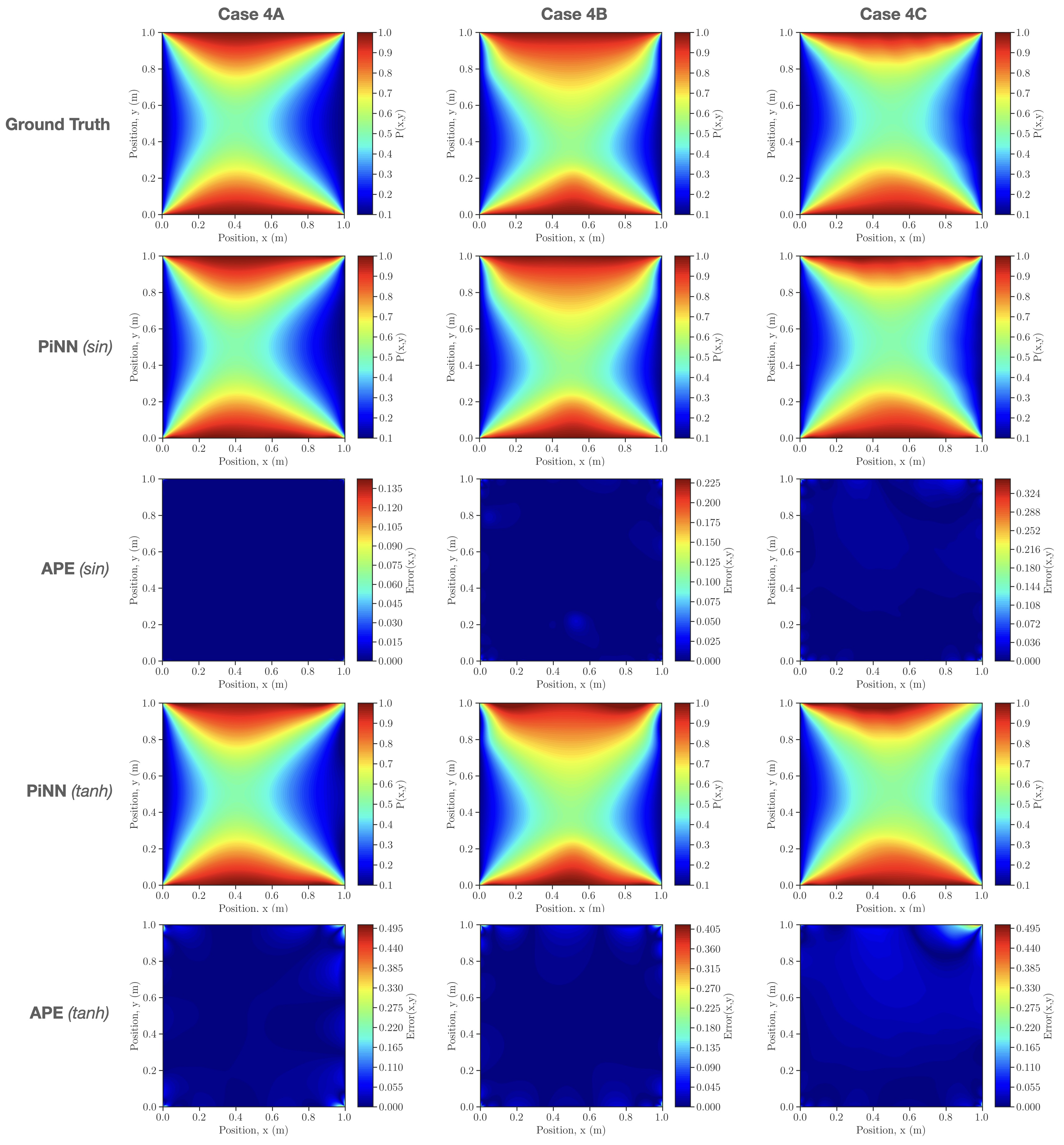}
    \caption{\textcolor{black}{A comparison of PiNN models using \textit{sin} and \textit{tanh} activation functions in predicting the pressure field in heterogeneous porous media, validated against the FEM solution with a $100\times100$ quadrilateral element grid. The comparison is illustrated for the pressure field considering $D_x = D_y = 0.02~m^2/s$, $\alpha = 0$, $\mu \phi = 0.001$, and $C_0 = 0.2~kg/m^3$. The absolute point error is also shown for each PiNN model to illustrate the mismatch between the prediction and ground truth solution. The PiNN model with the \textit{tanh} activation function encounters more difficulty in converging to the ground truth solution as the variation of permeability in the domain increases (i.e., as the length-scale of structural features decreases). In contrast, this issue is less pronounced for the PiNN model with \textit{sin} activation function.  }}
    \label{fig:Pres_results}
\end{figure}

\begin{figure}[H]
\centering
\includegraphics[width=0.99\linewidth]{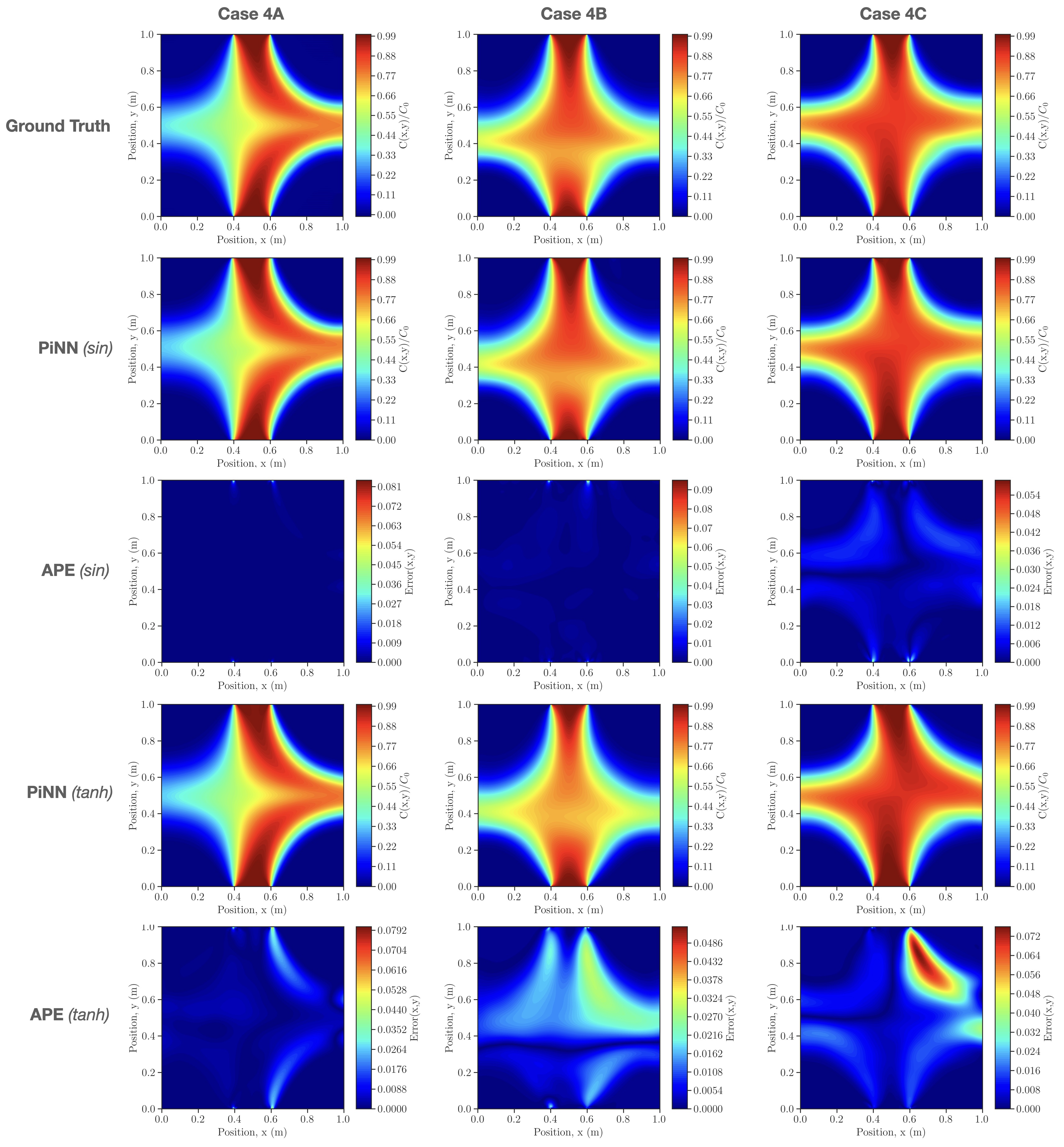}
    \caption{\textcolor{black}{A comparison of PiNN models using \textit{sin} and \textit{tanh} activation functions to predict the concentration field in heterogeneous porous media, validated against the FEM solution with a $100\times100$ quadrilateral element grid. The comparison is illustrated for the concentration  field  at $t = 1.00~s$ considering $D_x = D_y = 0.02~m^2/s$, $\alpha = 0$, $\mu \phi = 0.001$, and $C_0 = 0.2~kg/m^3$. The absolute point error is also shown for each PiNN model to illustrate the mismatch between the predictions and ground truth solutions. The PiNN model with  \textit{tanh} activation function encounters more difficulty in converging to the ground truth solution as the variation of permeability in the domain increases  (from left to right). This disparity is attributed to the rapid variation of the velocity field caused by heterogeneities 
    (i.e., as the length-scale of structural features). In contrast, the mismatch is less pronounced for the PiNN model with \textit{sin} activation function, making it two orders of magnitude more accurate than the PiNN model with \textit{tanh} activation function.}}
    \label{fig:con_results}
\end{figure}

Two PiNN models with five hidden layers and \{32,~32,~16,~16,~16\} neurons per layer are trained using   \textit{sin} and \textit{tanh} activation functions, respectively, to resolve the solute transport in these three heterogeneous cases  (i.e., predict the pressure and concentration fields). For each case, due to changes in the permeability field, the hyperparameter tuning process leads to different randomly distributed collocation points. For Case 4A, a total of 30,000 spatiotemporal random points (15,000 points within $0\leq x, ~y \leq1, ~0 \leq t \leq 1$ and 15,000  points on the boundaries) are used as collocation points to enforce the PDE, initial, and boundary loss terms. For Case 4B and Case 4C, the hyperparameter tuning process leads to a total of 36,000 and 40,000 spatiotemporal random points as collocation points, respectively. The weights for  the loss terms are also tuned for both PiNN models to achieve solution convergence while meeting the boundary condition requirements satisfactorily. Table \ref{tab:my-tableP} reports the collocation points within the domain and the weights of the loss terms used to predict the pressure and concentration fields. It also reports the comparison of the PiNN models, in terms of accuracy and training time, to predict the ground truth solution. Figures \ref{fig:Pres_results} and \ref{fig:con_results} illustrate the comparison between the PiNNs' predictions with the ground truth (FEM solutions) obtained for solute transport in heterogeneous porous media. The comparisons are shown for the steady state pressure field (Fig.~\ref{fig:Pres_results}) and the concentration field at $t=1.00~s$ (Fig.~\ref{fig:con_results}).

\begin{table}[H]
\caption{\textcolor{black}{A comparison of PiNN models using \textit{sin} and \textit{tanh} activation functions to predict the pressure and concentration fields in heterogeneous porous media, validated against the FEM solution with a $100\times100$ quadrilateral element grid. The MSE is calculated in (x,y) domain using Eq.~\ref{eq: MSE} for the pressure field at steady state and for the concentration field at $t = 1.00~s$. The training time of the PiNNs is measured through three runs on a system equipped with a 3.00-GHz 48-core Intel Xeon Gold 6248R CPU, Nvidia Quadro RTX 8000 GPU, and 128 GB of RAM. Additionally, the number of collection points in the domain and the weights assigned to the loss terms are reported for each PiNN model.} }
\label{tab:my-tableP}
\begin{tabular}{p{3.2cm}p{1.8cm}p{1.8cm}p{1.8cm}p{1.8cm}p{1.8cm}p{1.8cm}}
\toprule
& \multicolumn{2}{l}{\textbf{Case 4A}} & \multicolumn{2}{l}{\textbf{Case 4B}} & \multicolumn{2}{l}{\textbf{Case 4C}}  \\   \cline{2-7}   \addlinespace
& \textit{sin}     & \textit{tanh}    & \textit{sin}     & \textit{tanh}   & \textit{sin}     & \textit{tanh}   \\ \hline
 \addlinespace
$N_p$ & 15000 & 15000 & 18000 & 18000 & 20000                & 20000                               \\  \addlinespace
$w_i$, $w_b$ & 1.0, 1.0 & 1.0,1.0 & 1.0, 1.0 & 1.0, 1.0  & 1.0, 1.0                & 1.0, 1.0                               \\  \addlinespace
$w^P_p$, $w^C_p$  & 0.20, 0.70 &  0.25, 0.65 &  0.15, 0.80 &   0.25, 0.70  &  0.15, 0.80                &  0.3, 0.90                               \\  \addlinespace
Pressure  MSE & \textbf{1.13e-05} & 1.38e-04  & \textbf{2.48e-05} & 1.96e-04 & \textbf{5.38e-05}                & 2.08e-04                                        
    \\ \addlinespace
Concentration MSE &\textbf{1.12e-06} &  1.10e-04 &   \textbf{3.62e-06} & 1.25e-04 &  \textbf{9.86e-06} &  2.72e-04                                               \\  \addlinespace
Training Time (s)     &  \boldsymbol{ $1345 \pm 112$} &  $2710 \pm 73$ &   \boldsymbol{ $1566 \pm 131$}  & $3038 \pm 92$ &  \boldsymbol{ $2042 \pm 179$}     &   $4113  \pm 143 $                  \\ 
\bottomrule
\end{tabular}
\end{table}

Figure \ref{fig:Pres_results} shows
a comparison of PiNN models using \textit{sin} and \textit{tanh} activation functions in predicting the pressure field in heterogeneous porous media represented by Case 4A, 4B, and 4C permeability fields (Fig.~\ref{perm}). The absolute point error for each case, compared to the ground truth obtained using  FEM  with $100\times100$ quadrilateral elements,  shows that the PiNN with the \textit{sin} activation function provides a more accurate prediction for the pressure field. Table \ref{tab:my-tableP} reports the comparison in terms of MSE (Eq.~\ref{eq: MSE}). Based on the MSE reported, the PiNN model with \textit{sin} activation function is an order of magnitude more accurate in predicting the pressure field. As can be seen in Fig.~\ref{fig:Pres_results}, the PiNN model with the \textit{tanh} activation function encounters more difficulty in converging to the ground truth solution as the variation of permeability in the domain increases (i.e., as the length-scale of structural features decreases). In contrast, this issue is less pronounced for the PiNN model with \textit{sin} activation function. 

\begin{figure}[H]
\centering
\includegraphics[width=0.999\linewidth]{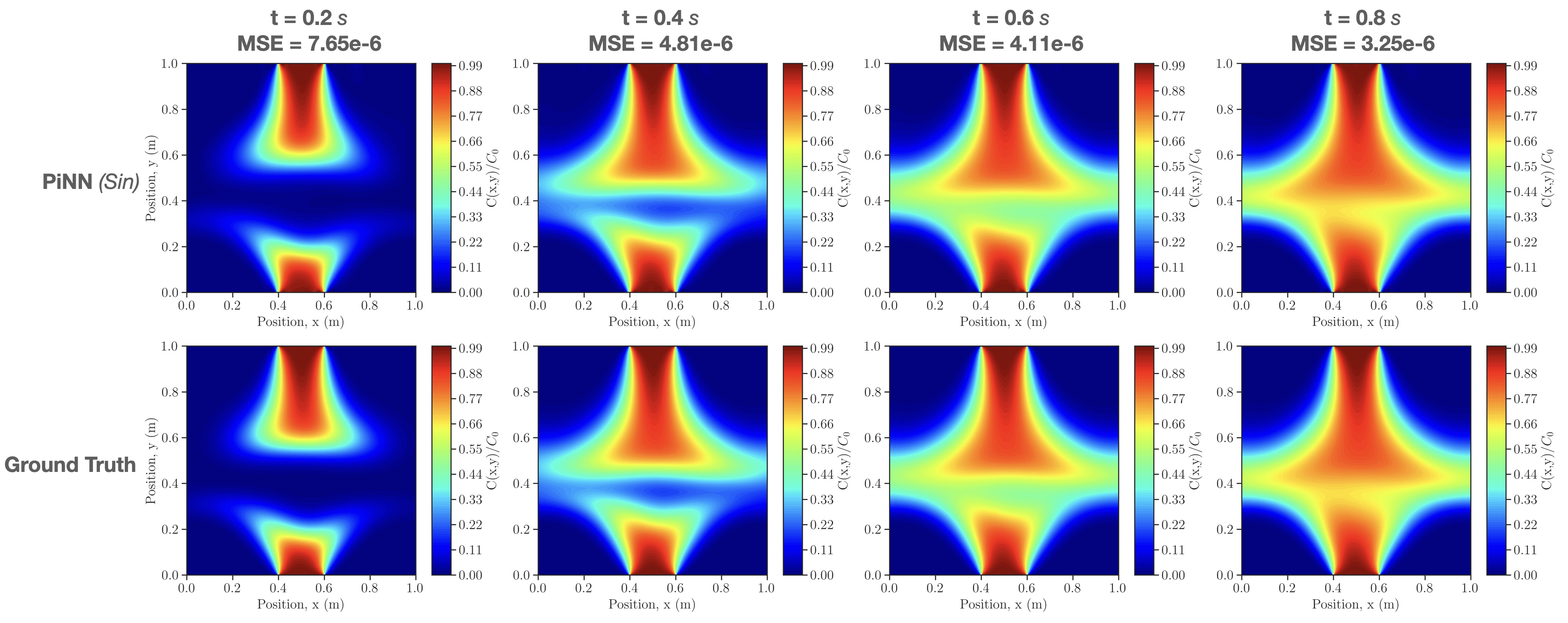}
    \caption{A comparison between the  PiNN's predictions (with \textit{sin} activation function) and ground truth solutions for transient solute transport in a heterogeneous porous medium with the given permeability field, Case 4B. The MSE is calculated using Eq.~\ref{eq: MSE} for the concentration fields obtained at each given time. }
    \label{fig:time_results}
\end{figure}

Figure \ref{fig:con_results} shows
a comparison of PiNN models using \textit{sin} and \textit{tanh} activation functions in predicting the concentration field in heterogeneous porous media represented by Case 4A, 4B, and 4C scaled permeability fields (Fig.~\ref{fig:perms}). The PiNNs' predictions are validated against the FEM solution with a $100\times100$ quadrilateral element grid. The comparison is illustrated for the concentration field at $t = 1.00~s$ considering $D_x = D_y = 0.02~m^2/s$, $\alpha = 0$, $\mu \phi = 0.001$, and $C_0 = 0.2~kg/m^3$. The absolute point error is also shown for each PiNN model to illustrate the mismatch between the prediction and ground truth solution. Table \ref{tab:my-tableP} reports the MSE  calculated using Eq.~\ref{eq: MSE}, and the training time. It is clearly observed that the mismatch for both PiNN models increases as  the length-scale of the structural features in the permeability field decreases (from left to right in Fig.~\ref{fig:con_results}), which is attributed to the rapid variation of the velocity field caused by heterogeneities. As shown, the PiNN model with  \textit{sin} activation function encounters less difficulty in converging to the ground truth solution, and this is more noticeable as the variation of permeability in the domain increases. In contrast, the mismatch is smaller for the PiNN model with  \textit{sin} activation function, making it two orders of magnitude more accurate and nearly 2x faster to train compared to the PiNN model using \textit{tanh} activation function. This order of magnitude for accuracy is consistent for all time steps. Figure \ref{fig:time_results} shows an example comparison between the  PiNN's predictions (with \textit{sin} activation function) and ground truth solutions for transient solute transport in a heterogeneous porous medium with a  permeability field described by Case 4B. The MSE is calculated using Eq.~\ref{eq: MSE} for the concentration fields obtained at each given time. These comparative findings clearly show that a PiNN with a periodic activation function can  solve for the pressure and solute concentration fields in 2D heterogeneous media with a higher degree of accuracy compared to \textit{tanh} activation function. The results also show that a PiNN with the periodic activation function reduces the training time. 

\begin{table}[H]
\centering
\caption {\textcolor{black}{A comparison of the inference time for the PiNN with  \textit{sin} activation function and FEM resolving the solute transport in the 2D homogeneous (Case 3) and heterogeneous (Case 4) porous media. The computational time is measured through  three runs on a system equipped with a 3.00-GHz 48-core Intel Xeon Gold 6248R CPU, Nvidia Quadro RTX 8000 GPU, and 128 GB of RAM. Note: This comparison only shows the inference speed-up factor, not considering the PiNN training time in the calculation. }} 
\label{Difference architectures}
\begin{tabularx}{\textwidth}{ >{\hsize=.4\hsize}X >{\hsize=0.2\hsize}X>{\hsize=0.2\hsize}X>{\hsize=0.2\hsize}X}
 \toprule  
 & {\textbf{FEM}} & {\textbf{PiNN}\textit{(sin)}} & {\textbf{Speed-up Factor} }   \\ 
  & {(s)} & {(s)} & {(—) }   \\ 
 \midrule

 2D Homogeneous  & $183.8 \pm 0.1$  & $0.126 \pm 0.01$ & 1458.7x   \\
 \addlinespace
 2D Heterogeneous - Case 4A & $363.2 \pm 0.3$  & $0.259 \pm 0.02$ & 1402.3x   \\
  \addlinespace
 2D Heterogeneous - Case 4B & $363.9 \pm 0.3$  & $0.260 \pm 0.01$ & 1399.6x   \\
  \addlinespace
 2D Heterogeneous - Case 4C & $364.5 \pm 0.2$  & $0.260 \pm 0.03$ & 1401.9x    \\
 
 \bottomrule
 \end{tabularx}\label{Table1}
\end{table}

Table \ref{Table1} compares the inference time for PiNN with the \textit{sin} activation function and FEM resolving solute transport in 2D homogeneous (case 3) and heterogeneous (Case 4) porous media. The speed-up factor achieved is around three orders of magnitude and can considerably escalate when a higher-resolution mesh (i.e., a higher number of elements) is used by the FEM solver; a situation that is  often faced in cases with higher levels of heterogeneities. In fact, in such cases, the number of collocation points used by PiNN also increases, which could negatively impact the cost of training but not the inference time. Therefore, the computational efficiency of PiNN is certainly superior to that of a traditional solver (e.g., FEM) in domains with complex geometry and  heterogeneity. It is also notable that PiNN comes with several drawbacks. The two most common drawbacks that were also faced in this study are: (i) PiNN's  training can face gradient vanishing problems, which can prohibitively slow down the learning process; and (ii) PiNN did not always converge due to competing non-linear loss terms (e.g., PDE loss terms related to the pressure and concentration fields in addition to the loss terms related to  initial/boundary conditions for both of the fields). It requires trials and errors or other adaptive techniques to adjust the loss terms' weight functions to mitigate the instability. The lack of a theoretical condition or constraint (e.g., Courant number in traditional computational fluid dynamics \cite{atmakidis2010study}) to ensure convergence is an open research area for investigation. Overall, considering the effectiveness of PiNN in combining scientific computing and DL and accelerating computation, the addition of PiNNs to the simulations of porous media flow should be further investigated. For example, a future study could be the assessment of PiNNs to model non-isothermal reactive flows with anisotropic dispersion coefficient in highly heterogeneous and deformable porous media.

\section{Conclusions}\label{sec:conclusion}
   
In this study, we demonstrated the use of physics-informed neural networks (PiNNs) and the advantages of employing a periodic activation function in addressing the solute transport problem in both homogeneous and heterogeneous porous media, as governed by the advection-dispersion equation. We constructed PiNNs using the \textit{sin} and \textit{tanh} activation functions to predict pressure and concentration fields within a given spatiotemporal domain. To evaluate the capabilities of PiNN models, we conducted seven case studies (1D and 2D) with varying degrees of permeability heterogeneity. We utilized an iterative random-search hyperparameter tuning method to determine the optimal architecture for each PiNN model in the test cases. The accuracy of the PiNNs' predictions was evaluated using absolute point error and mean square error metrics, and compared to the ground truth solutions obtained analytically or through FEM numerical methods. Our results showed that the PiNN with \textit{sin} activation function was capable of accurately predicting the behavior of the solute transport under a variety of conditions. The PiNN model employing the \textit{sin} activation function was found to be up to two orders of magnitude more accurate and up to two times faster to train than the commonly used \textit{tanh} activation function when applied to 2D homogeneous and heterogeneous porous media exhibiting structural features with distinct length scales relative to the domain size. Furthermore, the inference speed-up results showed that the PiNN's simultaneous predictions of pressure and concentration fields can reduce computational time by three orders of magnitude compared to FEM simulations.

\color{black}

\section{Acknowledgements}
S.A.F. would like to acknowledge the support by the Texas Sate University's International Research Accelerator Grant  (award no. 9000003039).

\section{Conflict of Interest}
The authors declare no conflict of interests.

\def\mybibdoicolor{\color{black}}
\newcommand*{\doi}[1]{\href{\detokenize{#1}} {\raggedright\mybibdoicolor{DOI: \detokenize{#1}}}}

\bibliographystyle{unsrtnat}
\bibliography{references.bib}
\end{document}